# Sideways Transliteration:
# How to Transliterate Multicultural Person Names?

**Raphael Cohen and Michael Elhadad**
**Ben-Gurion University of the Negev**
raficohen17@gmail.com

**Abstract**

In a global setting, texts contain transliterated names from many cultural origins. Correct transliteration depends not only on target and source languages but also, on the source language of the name. We introduce a novel methodology for transliteration of names originating in different languages using only monolingual resources. Our method is based on a step of noisy transliteration and then ranking of the results based on origin specific letter models. The transliteration table used for noisy generation is learned in an unsupervised manner for each possible origin language. We present a solution for gathering monolingual training data used by our method by mining of social media sites such as Facebook and Wikipedia. We present results in the context of transliterating from English to Hebrew and provide an online web service for transliteration from English to Hebrew[1].

## 1. Introduction

A common approach for translation of proper names is transliteration, spelling the word in the target language based on phonetics. News and business texts in English may contain transliterated named entities from any possible language.

For example, the news report sample sentence "UK's Minister for Africa *Henry Bellingham* met with *Alhaji Muhammad*", contains a name of English origin and a name of Arabic origin. The problem of translating this sentence to Arabic requires backwards transliteration for the name "*Alhaji Muhammad*" and forward transliteration for "*Henry Bellingham*."

The tasks of backward and forward transliteration were addressed by (Haizhou et al. 2004; Knight & Graehl 1998; Oh & Choi 2000; Stalls & Knight 1998; Virga & Khudanpur 2003; Wan & Verspoor 1998) suggesting a mix of methods based on pronunciation dictionaries for forward transliteration where available, learning transliteration rules from large lists of transliterated pairs, using monolingual named entities lists to produce possible transliterations in the target language and use of parallel text to acquire possible transliterations.

Backward transliteration assumes that only source and target languages are relevant for the transliteration process. When the proper name originates from a language different from both source and target, however, the result in the target language may be different. For example, when translating to Hebrew the example above, the "H" in "*Henry*" originates from the English "HH" sound, in contrast the "H" in "*Alhaji*" originates from Arabic and should be translated using the guttural ה ("h") of Hebrew. This case of transliterating an Arabic name transliterated in English back to Hebrew falls between the definitions of backward and forward transliteration. This setting appears to be quite frequent: (Al-Onaizan & Knight 2002b) report that 48% and 21% of the names in their two test sets were from "other language" origin. Reported accuracy for these names was significantly lower than for names of English origin (decrease of 20%-50%).

Li *et al.* (2007) described a method for transliterating from English to Chinese using knowledge of a name's language of origin (English, Chinese or Japanese) to provide a prior in a Bayes model. The training data used was bilingual name lists (English-Chinese) comprising 42K English origin name pairs, 83K Japanese origin pairs and 2M Chinese origin pairs.

---
[1] URL to resources withheld for review.

Khapra and Bhattacharyya (Khapra & Bhattacharyya 2009) suggested a preprocessing stage of origin detection (English, Hindi or Kannada) coupled with lexical lookup in post processing. Training pairs were extracted from a parallel corpus using Giza++(Och & Ney 2000).

Irvine *et al. (2010)* suggested a method for extracting training pairs from Wikipedia for multiple languages. However, their method does not take origin into account. This may cause errors such as: "Henri" in French origin name "Henri Charrier" is transliterated with "אנרי" while transliteration of the name "Henry Ian Cusick" of English origin would be "הנרי".

In this paper, we describe an approach for transliterating person names taking into consideration the names origins; we discuss the distinction between forward and backward transliteration to transliteration of words from an origin foreign to both source and target languages. We present a method that does not require the assembly of expansive parallel resources and is based solely on mono-lingual resources easily produced by native speakers and mined from social media resources such as Facebook. Different transliteration tables are learned in an unsupervised manner for different origins, requiring resources only in the target language. We use a phonetic lexical lookup scheme in order to look for known forms of spelling of the transliterated name. For names of Arabic origin phonetic lexical lookup improves accuracy by 15% from 58% to 67%, the unsupervised step of transliteration table adjustment further improves the accuracy to 84%. In total an improvement of 44%. For Hebrew, the same unsupervised method improved accuracy by 17% in total from 65% to 76%.

The rest of the paper is organized as follows. In Section 2, we describe current methods for transliteration and the required resources for each method. Section 3 presents our model method of transliterating from English to Hebrew. Section 4 discusses how the transliteration process should be expanded. In section 5, we suggest methods for origin identification of names.

## 2. Related Work

Knight and Graehl (1998) approached the problem of back transliteration using the CMU pronunciation dictionary to transform English words into phonetics, these phonetics transcriptions were transformed into the equivalent Japanese phonetics and various spellings in the Japanese transliteration script, Katanaka, were produced. The probabilities for the transition lattice were learned using EM on an English-Katanaka dictionary with 8,000 pairs.

Al-Onaizan and Knight (2002a) suggested a model independent of the existing CMU pronunciation based on spelling for transliteration from Arabic to English. Accuracy of transliterations of non-Arabic proper names from languages other than English was 20%-40% lower than backward transliteration of names originating in English.

Lin and Chen (2002) introduced a method for Chinese to English backwards transliteration based on transformation of both English words and transliterated Chinese words into phonetics and learning a probabilistic model of when two Chinese-English phonetic expressions are similar.

Al-Onaizan and Knight (Al-Onaizan & Knight 2002b) suggested an improved scoring method for backward transliterations using Web and corpus search.

Zhou et al. (2008) combine HMM classifiers trained on bigram and trigram letter models for scoring transliterations with Web mining of names frequency. Their method boosts the performance of an English-Arabic back transliteration.

Li *et al.* (2007) use EM to learn transliteration probability (token s in English is translated into token t in Chinese given the previous tokens). They showed that different characters are used when transliterating names of Japanese origin written in English than native English names. The language of origin for each name is incorporated by training a different model for each origin (English, Chinese and Japanese), the models are incorporated via a general transliteration model to reduce errors created by wrong origin detection. This method improved accuracy by 25.8% in average. The training data used was bilingual name lists (English-Chinese) comprising 42K English origin name pairs, 83K Japanese origin pairs and 2M Chinese origin pairs.

Khapra and Bhattacharyya (2009) use CRF to learn transliteration probabilities of one-grams, bi-grams and trigrams. Two separate models are trained, with a distinction between names of Indic origin (Hindi and Kannada) and English origin. They introduced a post-processing stage of lexicon lookup in order to find commonly used forms. Total improvement to accuracy was 7.1%. The size of the training resource was not reported, however, it requires a parallel corpus as pairs are extracted with Giza++.

Bhargava and Kondrak (2010) suggested a method for name origin detection using SVM classifier trained on ngram counts to improve transliteration from English to Hindi.

## 3. English to Hebrew Transliteration

Our method is developed as part of a transliteration system for person names from English to Hebrew. Previous work addressed the Hebrew to English direction (Kirschenbaum & Wintner 2009).

### 3.1 Forward Transliteration

For *Forward Transliteration* (transliterating a name of English origin to Hebrew) we use 3 steps:
a) The English name is transformed to a phonetic representation using the CMU Logios text to phonetics service[2].
b) We automatically generate many possible transliterations (most of them are likely to be inaccurate or noisy). This is done by consulting a manually crafted phonetics transliteration table which maps each phoneme to a list of possible Hebrew orthographic forms. All the possible representations are combined to generate the list of possible transliterations.
c) We then rank the noisy transliterations list using a combination of character-level language models. The score for each transliterated token is calculated by combining probabilities based on a language characters n-gram model: $M_1..M_4$, which are uni-grams, bi-grams, tri-grams and 4-grams. The probabilities are combined linearly where: $$P(w | Mi) = \sum_{i \in 1..4} 0.25 P(w | M_i)$$

This model was suggested by Goldberg and Elhadad (2008).The N-gram model is learned from a list English names transliterated in Hebrew (see Sec.5).

### 3.2 Backward Transliteration

*Backward Transliteration* is somewhat more complicated as there is only one correct match for the name in Hebrew. Using the aforementioned phonetics prediction for names of Hebrew origin produces poor results at best (i.e., the correct transliteration may not be in the generated list at all). Some consonants such as gutturals (ע,ח) and frequent sequences of two vowels such as "ai" ("*Yair*") "oi" ("*Roi*") or "ae" ("*Yael*") are missing from the English pronunciation. (They are pronounced as two syllables in Hebrew – *e.g.,* "o-i" and not as "oy").

A naïve solution for this problem is manually crafting a different letter based transliteration table for Hebrew. Given such a table and two monolingual lists of names: 1) Latin script names of Hebrew origin and 2) Hebrew script names of Hebrew origin (see Section 5 for our method to obtain these lists) we used the following method:
a) A noisy list of possible transliterations is generated: each Latin letter is replaced by possible Hebrew orthographic forms by consulting the spelling based transliteration table.
b) Top 2 transliterations are chosen based on the n-gram model described in Section 3.1 trained on the monolingual Hebrew names list.
c) The two transliterations are used for a lexicon lookup using edit distance with a phonetic based substitution table. The table is based on classic phonetics allowing replacement with distance of less than one to letter of the same semantic class (For example, Labial consonants: "bvmp", "בומפ"). If a match is found it replaces the noisily produced token.

| Roman script | Hebrew | Roman script | Hebrew |
|---|---|---|---|
| a | א,ע,ε | *a | ע,א |
| b | ב | c | ש,ק,כ,ח,צ |
| ch | ט',כ,צ,ח | ck | ק |
| d | ד | e | א,י,א, ε |
| e* | ה | f | פ |
| g | ג | ph | פ |
| h | ה,ח, ε | i | י, ε |
| *i | אי | K | כ,ק |
| j | ג,ג' | n | נ |
| m | מ | o | או,ו, ε |
| *o | עו,או | p | פ |
| q | ק | s | ש,ס |

---

| | | | |
|---|---|---|---|
| r | ר | t | ט,ת |
| sh | ש | w | ו,יו,וו |
| tz | צ | u | ו,יו |
| v | ב,ו | x | קס |
| y | י | z | ז,צ |
| a'a | ע | ae | א,ע |
| aa | ע | oe | וע,וא |
| oi | וי,עי | ui | י |
| eu | או,עו | aw | או |

**Table 1** Manually crafted spelling based transliterations table for Backwards Transliteration.[3]

## 4. Sideways Transliteration

### 4.1 Error Analysis

While our method produced acceptable results for person names of English and Hebrew origin names, results for names of Arabic origin transliterated were rarely accurate to the way these names are transliterated in Hebrew (see Table 2).

The same errors occur with names from other origins (Spanish, Turkish, Chinese, etc. See Table 2).

| English | Correct Hebrew form | Top 2 using English model |
|---|---|---|
| Rafael | רפאל | ראפייל,רפייל |
| Eitan | איתן | איטן,איטון |
| Haim | חיים | היים,האים |
| Nachman | נחמן | נקמן,נקמון |
| Azzam | עזאם | אסם,אסום |
| Omar | עומר/עמר | אומר,אומאר |
| Alatawna | אלאתונה/אלטונה | אלטנה,אלטונה |
| Saddam | סדאם | סדם,סודם |
| Hussein | חוסיין | הוסין, הוסיין |
| Juan | חואן | וון,וואן |
| Ruiz | רויז | רוויס,רוויז |

**Table 2** Transliteration error analysis – Note the poor results if using English n-grams model for transliterations, the 4 first names are from Hebrew origin; the rest are Arabic and Spanish.

### 4.2 Problem Definition

We refer by *Sideways Transliteration* to the generative process of transliterating names while considering the origin of the word and the pronunciation similarities of the target language and the name's origin. For example, when transliterating the name "Juan" to Hebrew, a translator takes into account the similarity between the pronunciation of the "J" sound in Spanish and the "ח" sound in Hebrew. This phenomenon is more acute when the target language and the origin of the name are more similar than the source language (Hebrew-Arabic vs. English). This problem would naturally occur when transliterating names of Hebrew origin from English to Arabic.

The marked difference between *Sideways Transliteration* to: *Backward* and *Forward Transliteration* is the many to many relation of the written forms: the nature of transliterations may create a variety of possible spelling of the name in both source and target languages (For example: we discovered 6 English transliterations of "محمد" : "Muhammed", "Mohamed", "Mhamed", etc. In Hebrew the name may be spelled in 2 ways: "מוחמד" and "מחמד"). In F*orward Transliteration*, an English origin name may have more than one correct spelling in Hebrew, and in *Backward Transliteration*, an Hebrew origin name has only one correct transliteration but possibly many forms in English.

### 4.3 Solution

We address the task of *Sideways Transliterations* by adding two steps to the transliteration process:

1. **Language Identification.** Classify each transliterated name according to its origin. In our system, we classify based on a list of frequent first names. This could be improved by incorporating a classifier based on n-grams of transliterated names in the source language (English in this case).
2. **Use of origin specific transliteration table.** Noisy production of all possible transliterations based on the origin specific table.
3. **Use origin specific N-gram Model.** Rank transliterations based on a model trained only on transliterated names of the same origin. If only a small number of training samples is available, the created model may be augmented by using transitions from the backward transliteration model. In our system, a Hebrew language n-gram model was constructed using 5,500 names, an Arabic model with 2,000 names and an English model with 2,000 names.
4. **Phonetic based lexical lookup.** Phonetics based lookup with edit distance (as described in Section 3.2). The phonetic substitution table is

---
[3] * marks the first or last letter.

```
(1)   transliterationTable ← uniformTable (not origin specific)
(2)   common ← tokens with 3 appearances or more
(3)   visited ← ∅
(3)   repeat
(4)     unmatched ← transliterate(transliterationTable, common), match to list
(5)     for token in Unmatched:
(6)       match,distance ← search for a match with minimal edit distance
(7)       if distance > ϵ:
(8)         continue
(9)       for $t_{ngram} \in token, m_{ngram} \in match$ :
(10)        candidate ← <$t_{ngram},m_{ngram}$>
(11)        transliterationTable ← transliterationTable ∪ {candidate}
(12)        if match ∈ transliterate(transliterationTable):
(13)          transitionsCount(candidate) ← transitionsCount(candidate) + 1
(14)        transliterationTable ← transliterationTable / {candidate}
(15)    transition ← most common transition in transitionsCount / visited
(16)    priorFit ← modelFit(transliterate(transliterationTable,common),model)
(17)    transliterationTable ← transliterationTable ∪ transition
(18)    posteriorFit ← modelFit(transliterate(transliterationTable,common),model)
(19)    if posteriorFit < priorFit:
(20)      transliterationTable ← transliterationTable / {transition}
(21)    visited ← visited ∪ {transition}
```

**Figure 1** - Learning algorithm for acquiring origin specific transitions from monolingual name lists. $\epsilon$ is the maximal distance cutoff, in our experiment it was initialized to 1. "model" is the trigram letter model over the target script names (of a specific origin).

maintained the same as it depends on the target language rather than on the origin of the transliterated name.

### 4.4 Unsupervised learning of origin specific transliteration table

Since our transliteration method (noisy generation + ranking) is unsupervised simply adding more possible transitions for all languages will add noise to the generation step.

Instead of manually crafting a specific spelling based transliteration table, as we did for *Backward Transliteration*. In Figure 1, we describe an iterative algorithm for acquiring the origin specific changes from a general transliteration table using the monolingual lists of names with the specific origin using only a subset of the common tokens (3 appearances or more) from both source and target name lists and a trigram letter model of the target name list.

The impact of new transitions that allow a match of a previously unmatched source token to one of the target tokens is tested using the letter model. Only transitions that improve the fit are added to the table to prevent noise.

Origin specific transliteration tables were learned for Arabic (see Table 3) and Hebrew (see Table 4).

| Roman Script | Hebrew Script | Improved fit |
|---|---|---|
| h | ח | True |
| a | ע | True |
| t | ט | True |
| y | י | True |
| m | מה | False |
| m | מג | False |
| s | סא | True |
| i | ימ | True |

**Table 3** – Transitions learned when iterating over the Hebrew origin name lists. Notice that the transitions: m -> מה, m -> מג were rejected by the language letter model.

| Roman Script | Hebrew Script | Improved fit |
|---|---|---|
| h | ח | True |
| a | ע | True |
| a | ׳ | True |
| d | ת | False |
| o | ות | True |
| tz | צי | True |
| tz | צ | True |
| m | מע | False |
| ll | ל | True |
| k | כז | False |
| i | ימ | True |
| i | יא | True |
| t | ט | True |

**Table 4** – Transitions learned when iterating over the Hebrew origin name lists. Transitions: d -> ת, m -> מע and k -> כז were rejected by the language letter model.

## 5. Acquiring Training Data using Facebook and Wikipedia

The training of extra classifiers and language models suggested in the previous section requires task specific training data:
1. For language identification, we require transliterations of names from the origin we are modeling.
2. For the target language model tuned to a specific origin, we require training sample in the target language.

Ideally, this can be solved by bilingual transliterated pairs dictionary of every language-pair modeled. However, constructing such resources would be expensive and for non-English language pairs this is not feasible.

### 5.1 Mining Facebook

With over six hundred million users worldwide and 75 supported languages, Facebook presents an excellent resource for extracting monolingual and bilingual data.

Users origins can be extracted by the information openly viewed in profiles (for some) and from group membership and connection to local personae in each country.

The international nature of Facebook encourages many users to transliterate their name in other languages, mainly English.

For this study we extracted 16,500 Hebrew names transliterated in English, 3,600 Arabic names transliterated in English and 2,000 Arabic names transliterated in Hebrew. All of these names were easily manually extracted from Facebook groups and 10 user profiles in less than 2 hours.

### 5.2 Mining Wikipedia

In recent years, Wikipedia has been a rich data source for NLP studies. Its multilingual nature is well suited for acquiring parallel and comparable texts in many pairs of languages (Erdmann et al. 2009; Kirschenbaum & Wintner 2010; Tyers & Pienaar 2008). Irvine *et al.* (Irvine et al. 2010) used the multilingual nature of Wikipedia pages to extract training data for transliteration. Their data did not contain origin information.

We used Wikipedia for extracting lists of names transliterated in Hebrew for our language n-gram model. The list feature of Wikipedia categories is especially useful for harvesting. For acquiring a language model for Sideways Transliteration from Spanish, we can use pages such as: "Spanish Artists", "Spanish National Football Team Players", "Argentina National Football Team Players", "Cities in Spain" and so on.

Many additional useful sources for English transliterations of names by origin can be found on the web. We used such lists for both common first names in Hebrew and Arabic.

## 6. Evaluation

We evaluated our method with two sets: a) names from our monolingual lists (used in the unsupervised training), the 250 most common Arabic origin tokens and 350 Hebrew origin tokens were sampled b) names extracted from news articles about the Middle East from CNN and BBC news. See Tables 3 and 4. An accurate result was a correct transliteration of the name, according to a native speaker, in one of the top 2 suggested names.

For the first set (names of Arab origin), baseline accuracy (transliteration without origin detection) was 58%. Lexicon lookup with phonetic edit distance improved accuracy to 67%. Unsupervised learning of transitions in the transliteration table further improved the accuracy to 84%, a total improvement of 44% in accuracy. For the second set (names of Hebrew origin), baseline accuracy (without the manually crafted table) was 65%, lexicon lookup with phonetic distance improved accuracy to 70.9% and unsupervised learning of the transliteration table further improved accuracy to 75%. For the same dataset, using the manually crafted table achieves recall of 84%.

For the third set, adding the *Sideways Transliteration* solution for Arabic improved accuracy for those names from 30.4% to 73.9% (Table 3).

| Origin | Proportion of test set | Accuracy |
|--------|------------------------|----------|
| English | 41% | 86.1% |
| Hebrew | 11.5% | 87.5% |
| Arabic | 26.5% | 73.9% |
| Other | 21% | 61% |

**Table 4** Accuracy of transliteration by origin of names.

## 7. Conclusions

The problem of *Sideways Transliteration* is present in international News Reports, see (Al-Onaizan & Knight 2002b), and is likely to occur in any geographic location with mixed ethnic populations such as transliteration into Indic languages as described by Khapra and Bhattacharyya (2009).

We have presented a method for transliterating person names from English to Hebrew, supporting both backward transliteration of Hebrew names and Sideways Transliteration of Arabic names. Average accuracy was 77.7%, with over 85% accuracy for names of English or Hebrew origin.

We suggested a method for improving transliteration of person names originating in languages other than the source or target languages in translation without requiring any bilingual resources. Our method relies solely on monolingual resources (*i.e.* name lists) and phonetic knowledge of the target language and is totally unsupervised.

The method of lexicon lookup with a phonetic substitution table outperforms simple lexicon lookup for transliterations, as the phonetic translation may not always be accurate. Using this method improved accuracy by 15% alone and is the basis of our unsupervised learning of an origin specific transliteration table.

Our solution improved accuracy for transliteration of names of Arabic origin by 44% for names extracted from Facebook and from 30.4% to 73.9% for names from news reports.

We suggested a simple method for leveraging data extracted from Wikipedia and Facebook in order to create monolingual resources required to implement sideways transliteration.